\documentclass[10pt,twocolumn,letterpaper]{article}

\usepackage{cvpr}
\usepackage{times}
\usepackage{epsfig}
\usepackage{graphicx}
\usepackage{amsmath}
\usepackage{amssymb,bm,booktabs,multirow,bigstrut,textcomp,subfigure,textcomp}

\usepackage[breaklinks=true,bookmarks=false]{hyperref}

\cvprfinalcopy 

\def\cvprPaperID{****} 

\begin{document}

\title{Deep Gradient Projection Networks for Pan-sharpening}

\author{Shuang Xu, Jiangshe Zhang\thanks{Corresponding author.}, Zixiang Zhao, Kai Sun, Junmin Liu, Chunxia Zhang\\
School of Mathematics and Statistics, Xi'an Jiaotong University, Xi’an 710049, China\\
{\tt\small \{shuangxu,zixiangzhao,\}@stu.xjtu.edu.cn, \{jszhang,kaisun,junminliu,cxzhang\}@mail.xjtu.edu.cn}
}

\maketitle

\begin{abstract}
   Pan-sharpening is an important technique for remote sensing imaging systems to obtain high resolution multispectral images. Recently, deep learning has become the most popular tool for pan-sharpening. This paper develops a model-based deep pan-sharpening approach. Specifically, two optimization problems regularized by the deep prior are formulated, and they are separately responsible for the generative models for panchromatic images and low resolution multispectral images. Then, the two problems are solved by a gradient projection algorithm, and the iterative steps are generalized into two network blocks. By alternatively stacking the two blocks, a novel network, called gradient projection based pan-sharpening neural network, is constructed. The experimental results on different kinds of satellite datasets demonstrate that the new network outperforms state-of-the-art methods both visually and quantitatively. The codes are available at \url{https://github.com/xsxjtu/GPPNN}.
\end{abstract}

\section{Introduction}

\begin{figure}
	\centering
	\includegraphics[width=\linewidth]{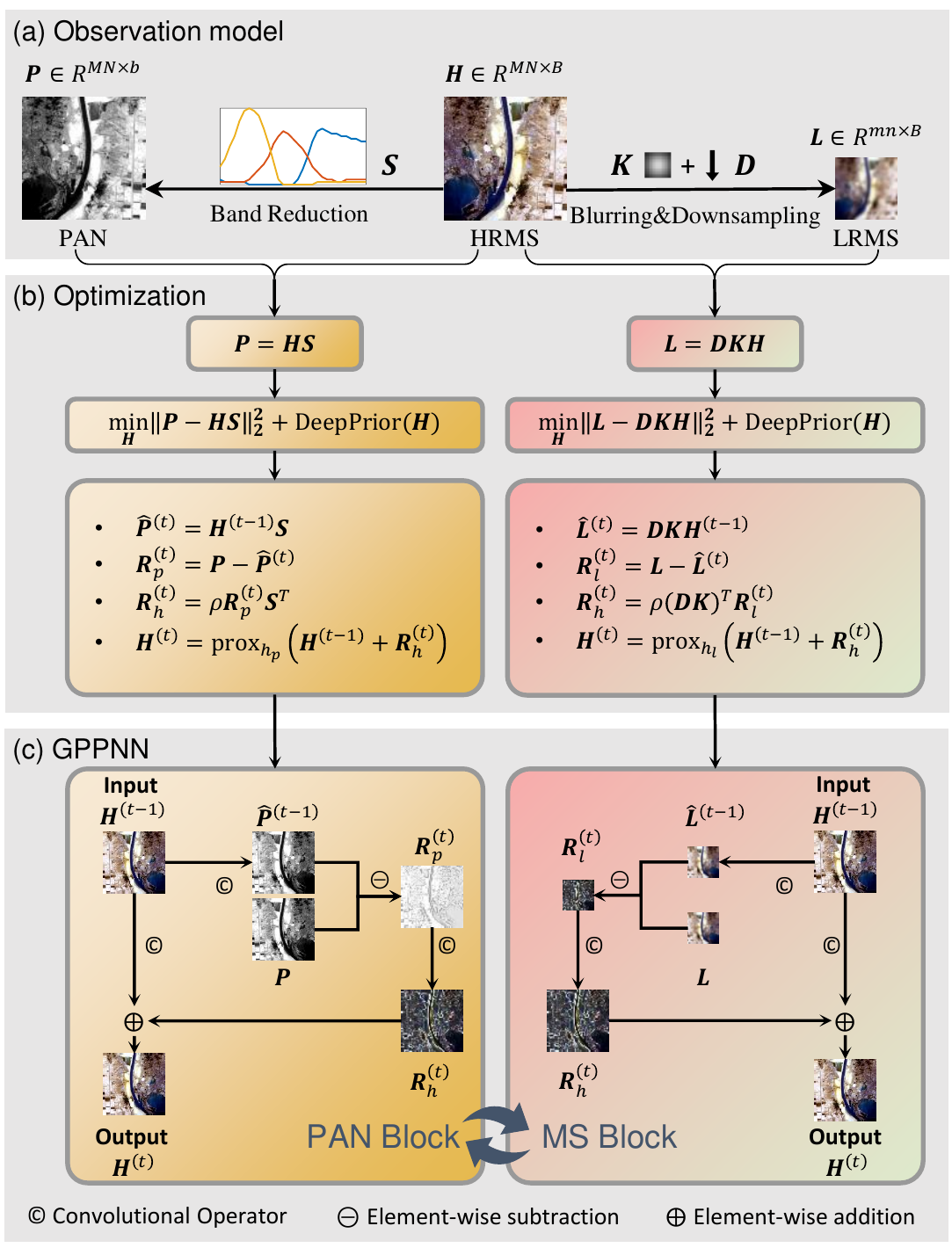}
	\caption{(a) The observation models for LRMS and PAN images. (b) Two formulated optimization problems and iterative steps of the gradient projection algorithm. (c) The main blocks in our proposed GPPNN. }
	\label{fig:model}
\end{figure}

Multispectral images store multiple images corresponding to each band (or say, channel) in an optical spectrum, and they are widely utilized in literature of remote sensing. With the limitation of imaging devices, satellites however often measure the low spatial resolution multispectral (LRMS) images \cite{8672156,9082183,6998089}. Compared with the multispectral image, the panchromatic (PAN) image is characterized by the high spatial resolution but only one band. Lots of satellites carry both multispectral and panchromatic sensors to simultaneously measure the complementary images, such as Landsat8, GaoFen2 and QuickBird. To obtain the high resolution multispectral (HRMS) image, a promising way is to fuse the complementary information of the LRMS image and the PAN image. This technique is called as {\it pan-sharpening} \cite{8672156,9082183}. 

Pan-sharpening can be cast as a typical image fusion on super-resolution problems. The past decades witnessed the development of pan-sharpening. The classic algorithms including component substitution (CS) \cite{IHS,GS}, multiresolution analysis (MRA) \cite{GLP,AWLP} and other techniques. In the era of deep learning, convolutional neural networks have emerged as a significant tool for pan-sharpening \cite{MA2019166}. One of the seminal work is the pan-sharpening neural network (PNN) proposed by Masi \etal \cite{PNN}. Borrowing the idea of the first super-resolution network \cite{SRCNN}, PNN is fed with the concatenation of a PAN image and an upsampled LRMS image to regress the HRMS image. 

In fact, there are only three convolutional units in PNN, so it is a relatively shallow network. Recently, numerous models have been proposed to improve the PNN. Owing to the triumphs of residual networks \cite{ResNet}, several papers utilize the shortcut or residual convolutional units to build deep networks, including MIPSM \cite{MIPSM}, DRPNN \cite{DRPNN} and PanNet \cite{PanNet}. They generally contain 10 or more convolutional units. Besides these networks, to make the best of advantages of neural networks, some researchers build deeper networks. For example, Wang \etal employ the densely connected convolutional unit \cite{DenseNet} to design a 44-layer network \cite{DensePanNet} for pan-sharpening. 

It is well-known that deepening the layers of networks does not necessarily improve the performance, since it is difficult to train deeper networks and redundant parameters make them easily over-fit. Very recently, the remote sensing community begins to rethink how to make the full use of PAN images' information \cite{8667040,9153037}. It is worthy noting that most the pan-sharpening networks regard the PAN image as a channel of the input. This manner ignores different characteristics between PAN and LRMS images. A growing number of researchers attempt to propose the two-branch networks \cite{LIU20201,MSDCNN}. In the first stage, the two branches separately extract the features for PAN and LRMS images. In the second stage, the features are fused to reconstruct the HRMS image. 

Although convolutional neural networks exhibit promising performance in pan-sharpening, they require a large amount of training samples \cite{Algorithm_Unrolling,Model-Meets-Deep-Learning}, and they do not account for the observation progress of PAN and LRMS images, i.e., lacking the interpretability. Therefore, there still leaves the room for improvement. The research on model-based deep learning is the trend in image processing field to close the gap between classic models and neural networks, and it is found that model-based deep networks usually outperform the intuitively designed networks \cite{Algorithm_Unrolling,Model-Meets-Deep-Learning}. Xie \etal present a multispectral and hyperspectral (HS) image fusion network (MHNet) for the {\it hyperspectral pan-sharpening} task \cite{MHNet}. There is no doubt that MHNet can be naturally adapted to pan-sharpening \cite{HAM-MFN}. Nonetheless, MHNet is designed to describe the low-rank property for hyperspectral images, and our experiments show that MHNet may perform badly in the pan-sharpening scenario.

In this paper, we develop a novel model-based deep network for pan-sharpening. Our contributions are summarized as follows:

Firstly, this paper considers the generative models for PAN and LRMS images. That is, as shown in Fig. \ref{fig:model}(a), PAN images are the linear combination of the bands in HRMS images, and LRMS images are generated by blurring and downsampling HRMS images. Combining the observation models and the deep prior, we propose two optimization problems, and they can be effectively solved by the gradient projection method as illustrated in Fig. \ref{fig:model}(b). 

Secondly, inspired by the idea of algorithm unrolling techniques, the iterative steps are generalized as two neural blocks separately justifying the generative models for PAN and LRMS images. The computational flows in the proposed neural blocks are interpretable. As show in Fig. \ref{fig:model}(c), for the MS Block, given a current estimation of the HRMS image, it generates corresponding LRMS image and computes the residual between the generated LRMS image and the real one. This residual then is upsampled and is added into the current estimation to reconstruct the next HRMS image. The PAN block can be interpreted similarly. We build a new network by alternatively stacking the two blocks. In what follows, it calls the gradient projection based pan-sharpening neural network (GPPNN). To the best of our knowledge, it is the first model-driven deep network for pan-sharpening. 

Thirdly, the proposed GPPNN is compared with the 13 state-of-the-art (SOTA) and classic pan-sharpening methods. The extensive experiments conducted on three popular satellites (i.e., Landsat8, QuickBird, GF2) demonstrate that our networks outperform other counterparts both quantitatively and visually. 

\section{Related work}
\subsection{Classic pan-sharpening methods}
The classic pan-sharpening methods mainly consists of CS based algorithms, MRA based algorithms and other algorithms. CS methods assume that the spatial and spectral information of a multispectral image can be decomposed. Therefore, an HRMS image is reconstructed by combining the spatial information of a PAN image and the spectral information of an LRMS image. In the past decades, researchers have  designed various decomposition algorithms. For example, intensity-hue-saturation (IHS) fusion \cite{IHS} employs the IHS transformation, Brovey method \cite{Brovey} uses a multiplicative injection scheme, and Gram-Schmidt (GS) method \cite{GS} exploits the Gram-Schmidt orthogonalization procedure. The main drawback of CS methods is that the image contains artifacts if the spectral and spatial information is not appropriately decomposed. The MRA methods apply the multi-scale transformation to PAN images to extract spatial details which then are injected into the upsampled LRMS images. Typical algorithms include high-pass filter (HPF) fusion \cite{HPF}, and Indusion method \cite{Indusion}, smoothing filter-based intensity modulation (SFIM) \cite{SFIM}. The performance of the MRA method strongly depends on the multi-scale transformation. 

\subsection{Deep learning based methods}
Recently, convolutional neural networks have been one of the most effective tools for remote sensing. Given a parameterized network, it is fed with an LRMS image and a PAN image to regress an HRMS image, and its parameters (or say, weights) are learned from data in the end-to-end fashion. The first attempt is the PNN with three convolutional units \cite{PNN}. Recently, thanks to the rapid development of computer vision \cite{ResNet,DenseNet}, it is able to train very deep networks. Researchers propose the deep pan-sharpening networks with dozens of layers and the performance has been greatly improved \cite{RSIFNN,DensePanNet,PanNet,MSDCNN}. At the same time, researchers also explore the two-branch networks to separately extract the features from MS and PAN images \cite{LIU20201,MSDCNN}. Recently, one of the research trends of the pan-sharpening community is to combine the classic methods with deep neural networks to improve the interpretability of the deep learning based methods. For example, inspired by the idea of MRA algorithms, MIPSM \cite{MIPSM} designs a spatial detail extraction network for the PAN images and injects the details into the LRMS images.  Liu \etal propose an adaptive weight network for integrating the advantages of different classic methods \cite{PWNet}. It overcomes the shortcomings of the CS and MRA algorithms, and outperforms some SOTA deep learning based methods.

\subsection{Model-driven deep networks}
Most of the deep neural networks are designed intuitively. Recently, a growing number of researchers focus on model-based neural networks for image processing tasks \cite{Algorithm_Unrolling,Model-Meets-Deep-Learning}. The basic idea of model-driven deep learning is to formulate an observation model or optimization problem by integrating the prior knowledge for a specific task and to translate each iteration of the algorithm step into a layer of deep neural networks \cite{Algorithm_Unrolling,Model-Meets-Deep-Learning}. Passing through the stacked layers corresponds to execute the algorithm with a certain number of times. Model-based deep learning builds the bridge between classic models and deep neural networks. This idea has been successfully applied in various tasks, including sparse coding \cite{LSC}, compressive sensing \cite{ADMM-CSNet}, image deblurring \cite{DUBLID}, image dehazing \cite{Proximal-Dehaze-Net} and image deraining \cite{RCDNet}. It is worth mentioning the MHNet, a model-driven network for the hyperspectral pan-sharpening task \cite{MHNet} to super-resolve HS images with the guidance of MS images. It can be naturally adapted to pan-sharpening, but MHNet mainly focuses on the low-rank property for HS images, i.e., its rank $r_{\rm HS}$ is far lower than the number of bands $B_{\rm HS}$. In practice, there are dozens or hundreds of bands in an HS image, while there are no more than 10 bands in an MS image. So, the low-rank property is not evident for MS images, and MHNet may break down in pan-sharpening task.

\section{GPPNN}
In this section, we develop a model-driven network for pan-sharpening. For convenience, we summarize the notations in this paper before presentation of the GPPNN. $\bm{L}\in R^{mn\times B}$ is an LRMS image with a height of $m$, a width of $n$ and the number of bands of $B$. $\bm{H}\in R^{MN\times B}$ is an HRMS image with a height of $M$, a width of $N$ and the number of bands of $B$. $\bm{P}\in R^{MN\times b}$ is a PAN image whose spatial resolution is the same with that of $\bm{H}$, but there is only one band (i.e., $b=1$). $r=M/m=N/n$ is the spatial resolution ratio.  With abuse of notations, we use their tensor versions in the context of deep learning (namely, $\mathcal{L}\in R^{m\times n\times B}, \mathcal{H}\in R^{M\times N\times B}, \mathcal{P}\in R^{M\times N\times b}$). Notation $\mathtt{Conv}(\cdot;c^{\rm in},c^{\rm out})$ is the convolutional operator whose input and output are with $c^{\rm in}$ and $c^{\rm out}$ channels, respectively. In what follows, the function $\mathtt{Conv}(\cdot;c^{\rm in},c^{\rm mid},c^{\rm out})$ denotes the cascaded convolutional operator, that is, 
\begin{equation}
\begin{aligned}
&\mathtt{Conv}\left(\cdot;c^{\rm in},c^{\rm mid},c^{\rm out}\right) \\
=& \mathtt{Conv}\left(\mathtt{ReLU}\left(\mathtt{Conv}\left(\cdot;c^{\rm in},c^{\rm mid}\right)\right);c^{\rm mid},c^{\rm out}\right).
\end{aligned}
\end{equation}

\subsection{Model formulation}\label{sec:model_formulation}
Our network starts with the observation model for the LRMS, HRMS and PAN images. It is assumed that an LRMS image is obtained by downsampling and blurring an HRMS image, while a PAN image is the result of spectral response for an HRMS image. In formula, we have $ \bm{L} = \bm{DKH}, \bm{P}=\bm{HS},$
where $\bm{D}\in R^{mn\times MN}$ denotes a downsampling matrix and $\bm{K}$ is the (low-passing) circular convolution matrix, and $\bm{S}\in R^{B\times b}$ is the so-called spectral response function. It is well-known that inferring the HRMS image is an ill-posed problem. Hence, it often formulates the following penalized optimization, 
\begin{equation}\label{eq:naive_model}
\min_{\bm{H}}  f(\bm{L},\bm{H})+g(\bm{P},\bm{H}) +\lambda h(\bm{H}),
\end{equation}
where $h(\cdot)$ is the prior term, and $f(\bm{L},\bm{H})=\|\bm{L}-\bm{DKH}\|_2^2/2$ and $g(\bm{P},\bm{H})=\|\bm{P}-\bm{HS}\|_2^2/2$ are data fidelity terms which are responsible to LRMS and PAN images, respectively. In the classic methods, $h(\cdot)$ is usually designed as a hand-craft function, such as the total variation or nuclear norm \cite{8167324}. However, in the era of deep learning, it is suggested to set $h(\cdot)$ as a deep prior \cite{Deep_Image_Prior,Deep_Denoiser_Prior}. In other words, it is better to set an implicit prior captured by the neural network parametrization. Additionally, the deep prior is learned from data and can adapt to different tasks and observation models. To make the best of deep prior, instead of the above issue, we consider an LRMS-aware problem and a PAN-aware problem:
\begin{subequations}\label{eq:problem}
	\begin{align}
	\min_{\bm{H}} & \frac{1}{2}\left\|\bm{L}-\bm{DKH}\right\|_2^2+\lambda h_{l}(\bm{H}), \label{eq:problem_LR}\\
	\min_{\bm{H}} & \frac{1}{2}\left\|\bm{P}-\bm{HS}\right\|_2^2+\lambda h_{p}(\bm{H}). \label{eq:problem_PAN}
	\end{align}
\end{subequations}
Here, $h_{l}(\cdot)$ and $h_{p}(\cdot)$ are two deep priors accounting for the observations of LRMS and PAN images, respectively. The ablation experiment in section \ref{sec:ablation_exp} verifies that Eq. (\ref{eq:problem}) achieves better results than Eq. (\ref{eq:naive_model}). In the next, we describe how to solve the two problems. Moreover, the solutions are generalized into an LRMS-aware block (MS Block) and a PAN-aware block (PAN block) that can be embedded into neural networks. 

\subsection{MS Block}\label{sec:LR_block}
We employ the gradient projection method \cite{Gradient_projection} to solve Eq. (\ref{eq:problem_LR}) and the updating rule is 
\begin{equation}\label{eq:gp_LR}
\bm{H}^{(t)} = \mathrm{prox}_{h_l} \left(\bm{H}^{(t-1)}-\rho \nabla f(\bm{H}^{(t-1)}) \right),
\end{equation}
where $\rho$ is the step size, $\mathrm{prox}_{h_l}(\cdot)$ is a proximal operator corresponding to penalty ${h_l}(\cdot)$ and $\nabla f(\bm{H}^{(t-1)})=-(\bm{DK})^T(\bm{L}-\bm{DKH})$ denotes the gradient of the data fidelity term. 

Inspired by the principle of model-driven deep learning \cite{Algorithm_Unrolling}, we generalize Eq. (\ref{eq:gp_LR}) as a network block. To begin with, Eq. (\ref{eq:gp_LR}) is split into four steps as follows,
\begin{subequations}
	\begin{align}
	\hat{\bm{L}}^{(t)} &= \bm{DK}\bm{H}^{(t-1)}, \label{eq:gp_LR1}\\
	\bm{R}_{l}^{(t)} &= \bm{L}-\hat{\bm{L}}^{(t)}, \label{eq:gp_LR2}\\
	\bm{R}_{h}^{(t)} &= \rho\left(\bm{DK}\right)^{T}\bm{R}_{l}^{(t)}, \label{eq:gp_LR3}\\
	\bm{H}^{(t)} &= \mathrm{prox}_{h_l}\left(\bm{H}^{(t-1)}+\bm{R}_{h}^{(t)}\right). \label{eq:gp_LR4}
	\end{align}
\end{subequations}

Then, each step is translated with deep learning terminologies. For convenience, we use the tensor versions to represent the variables in the context of deep learning. In Eq. (\ref{eq:gp_LR1}), given a current HRMS image $\mathcal{H}^{(t-1)}$, it generates an LRMS image $\hat{\mathcal{L}}^{(t)}$ by applying a low-passing filter and downsampling. In neural networks, this step is implemented by 
\begin{equation}\label{eq:nn_LR1}
\hat{\mathcal{L}}^{(t)} = \mathtt{Conv}\left(\mathcal{H}^{(t-1)}; B,C,B\right) \downarrow,
\end{equation} 
where downsampling is conducted with a bicubic interpolation $\downarrow$ and the filter $\bm{K}$ is replaced by a cascaded convolutional operator $\mathtt{Conv}(\cdot; B,C,B)$ to obtain more expressive features. $C$ is the number of channels for the feature maps, and we set it to 64 in this paper. $B$ is the number of channels for MS images, and it depends on the input data. 

Afterwards, Eq. (\ref{eq:gp_LR2}) computes residuals between the real LRMS image $\mathcal{L}$ and the generated LRMS image $\hat{\mathcal{L}}^{(t)}$, and the translation is trivial as shown in following equation, 
\begin{equation}\label{eq:nn_LR2}
\mathcal{R}_{l}^{(t)} = \mathcal{L}-\hat{\mathcal{L}}^{(t)}.
\end{equation}
In the next, Eq. (\ref{eq:gp_LR3}) obtains the high-resolution residuals. Analogous to Eqs. (\ref{eq:gp_LR1}) and (\ref{eq:nn_LR1}), this step is rewritten as 
\begin{equation}\label{eq:nn_LR3}
\mathcal{R}_{h}^{(t)} = \rho\mathtt{Conv}\left(\mathcal{R}_{l}^{(t)}; B,C,B\right)\uparrow.
\end{equation}
Remark that the filters in Eqs. (\ref{eq:gp_LR1}) and (\ref{eq:gp_LR3}) transpose to each other, but we do not force the convolutional kernels in Eqs. (\ref{eq:nn_LR1}) and (\ref{eq:nn_LR3}) to satisfy this requirement for flexibility. The ablation experiment in section \ref{sec:ablation_exp} shows that it slightly improves GPPNN's performance. At last, Eq. (\ref{eq:gp_LR4}) outputs the HRMS image by taking the residual into account with a proximal operator. As illustrated before, proximal operators regarding the deep prior are modeled by the deep networks \cite{Deep_Image_Prior,Deep_Denoiser_Prior}. In this manner, the deep prior can be learned implicitly from data. So, we have
\begin{equation}\label{eq:nn_LR4}
\mathcal{H}^{(t)} = \mathtt{Conv}\left(\mathcal{H}^{(t-1)} + \mathcal{R}^{(t)}_{h}; B,C,B\right).
\end{equation}
In what follows, Eqs. (\ref{eq:nn_LR1}), (\ref{eq:nn_LR2}), (\ref{eq:nn_LR3}) and (\ref{eq:nn_LR4}) are named as an MS Block. For better understanding, the computational flow for an MS Block is displayed in Fig. \ref{fig:model}(c). 

\begin{figure*}
	\centering
	\includegraphics[width=.8\linewidth]{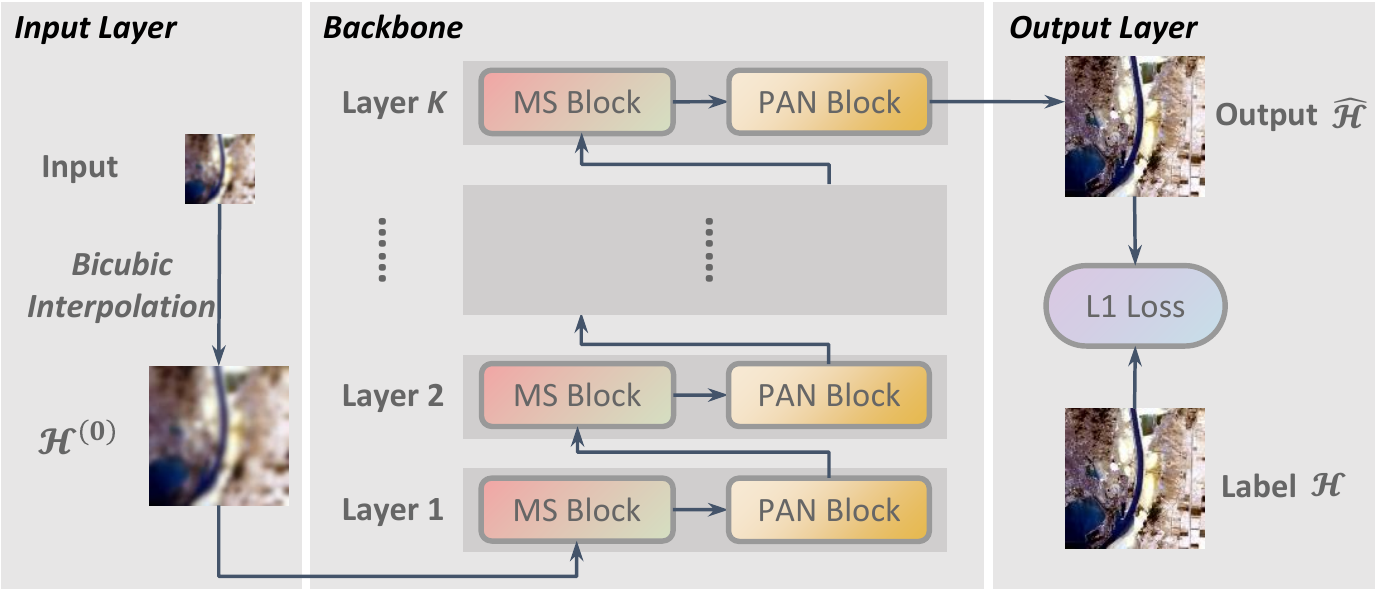}
	\caption{The structure of GPPNN.}
	\label{fig:structure}
\end{figure*}

\subsection{PAN block}\label{sec:PAN_block}
In this subsection, we consider the observation model for PAN (i.e., Eq. (\ref{eq:problem_PAN})). With the gradient projection method, the updating rule is
\begin{equation}\label{eq:gp_PAN}
\bm{H}^{(t)} = \mathrm{prox}_{h_p} \left(\bm{H}^{(t-1)}-\rho \nabla g(\bm{H}^{(t-1)}) \right),
\end{equation} 
where $\nabla g(\bm{H}^{(t-1)})=-(\bm{P}-\bm{HS})\bm{S}^{T}$. With the similar techniques, it is able to translate Eq. (\ref{eq:gp_PAN}) into a block of neural networks. At first, Eq. (\ref{eq:gp_PAN}) is split into four steps as follows,
\begin{subequations}
	\begin{align}
	\hat{\bm{P}}^{(t)} &= \bm{H}^{(t-1)}\bm{S}, \label{eq:gp_PAN1}\\
	\bm{R}_{p}^{(t)} &= \bm{P}-\hat{\bm{P}}^{(t)}, \label{eq:gp_PAN2}\\
	\bm{R}_{h}^{(t)} &= \rho\bm{R}_{p}^{(t)}\bm{S}^T, \label{eq:gp_PAN3}\\
	\bm{H}^{(t)} &= \mathrm{prox}_{h_p}\left(\bm{H}^{(t-1)}+\bm{R}_{h}^{(t)}\right). \label{eq:gp_PAN4}
	\end{align}
\end{subequations}
In the context of deep learning, as shown in Fig. \ref{fig:model}(c), these steps are rewritten as, 
\begin{subequations}\label{eq:nn_PAN}
	\begin{align}
	\hat{\mathcal{P}}^{(t)} &= \mathtt{Conv}\left(\mathcal{H}^{(t-1)}; B,C,b\right), \label{eq:nn_PAN1}\\
	\mathcal{R}_{p}^{(t)} &= \mathcal{P}-\hat{\mathcal{P}}^{(t)}, \label{eq:nn_PAN2}\\
	\mathcal{R}_{h}^{(t)} &= \rho\mathtt{Conv}\left(\mathcal{R}_{p}^{(t)}; b,C,B\right), \label{eq:nn_PAN3}\\
	\mathcal{H}^{(t)} &= \mathtt{Conv}\left(\mathcal{H}^{(t-1)} + \mathcal{R}^{(t)}_{h}; B,C,B\right). \label{eq:nn_PAN4}
	\end{align}
\end{subequations}
Here, $b=1$ is the number of channel for PAN images. Remark that the underlying assumption of Eq. (\ref{eq:problem_PAN}) is that the PAN image is a linear combination of the HRMS image. $\bm{S}$/$\bm{S}^T$ is regarded as a band reduction/expansion operator. With this assumption, convolutional units in Eqs. (\ref{eq:nn_PAN1}) and (\ref{eq:nn_PAN3}) should be with the kernel size of 1.  

\subsection{GPPNN}\label{sec:GPPNN}

Now, with the MS Block and the PAN block, we are ready to construct the gradient projection based pan-sharpening neural network (GPPNN). The structure of our GPPNN is shown in Fig. \ref{fig:structure}. The network starts with an input layer, and it requires an initial value of the HRMS image. We initialize $\mathcal{H}^{(0)}\in R^{M\times N\times B}$ by applying the bicubic interpolation to the input LRMS image $\mathcal{L}\in R^{m\times n\times B}$. The network is followed by a backbone subnetwork. There are $K$ layers, each of which consists of an MS Block and a PAN block. In formula, there are 
\begin{equation}\label{eq:nn_LR_Block}
\mathcal{H}^{(t+0.5)} = \mathtt{MS\_Block}\left(\mathcal{H}^{(t)}, \mathcal{L}\right) 
\end{equation}
and
\begin{equation}\label{eq:nn_PAN_Block}
\mathcal{H}^{(t+1)} = \mathtt{PAN\_Block}\left(\mathcal{H}^{(t+0.5)}, \mathcal{P}\right).
\end{equation}
The output of the last layer, denoted by $\hat{\mathcal{H}}\in R^{M\times N\times B}$, is the final reconstructed HRMS. 

\subsection{Training details}
Our GPPNN is supervised by the $\ell_1$ loss between $\hat{\mathcal{H}}$ and the ground truth $\mathcal{H}$, 
$
\|\hat{\mathcal{H}}-\mathcal{H}\|_1.
$
The paired training samples are unavailable in practice. When we construct the training set, the Wald protocol \cite{Wald97fusionof} is employed to generate the paired samples. For example, given the multispectral image $\mathcal{H}\in R^{M\times N\times B}$ and the PAN image $\mathcal{P}\in R^{rM\times rN\times b}$, both of them are downsampled with ratio $r$, and the downsampled versions are denoted by $\mathcal{L}\in R^{M/r\times N/r\times B}$ and $\tilde{\mathcal{P}}\in R^{M\times N\times b}$. In the training set, $\mathcal{L}$ and $\tilde{\mathcal{P}}$ are regarded as the inputs, while $\mathcal{H}$ is the ground-truth. 

GPPNN is implemented with Pytorch framework. They are optimized by Adam over 100 epochs with a learning rate of $5\times 10^{-4}$ and a batch size of 16. In our experiments, $k_{\rm LR}=3$ and $k_{\rm PAN}=1$. In section \ref{sec:K_C}, we report the performance of GPPNN with different $C$ and $K$. As a balance, $C$ and $K$ are set to 64 and 8, respectively.  

\begin{figure*}[h]
	\centering
	\subfigure[ {\scriptsize LRMS}] {\includegraphics[width=.15\linewidth]{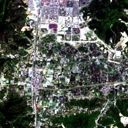}}  
	\subfigure[ {\scriptsize PAN}] {\includegraphics[width=.15\linewidth]{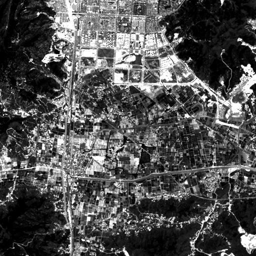}}  
	\subfigure[ {\scriptsize GT(PSNR)}] {\includegraphics[width=.15\linewidth]{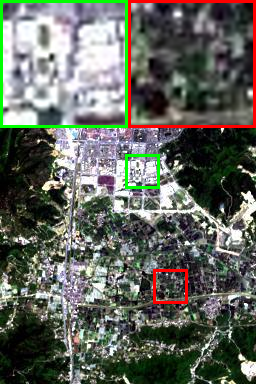}}  
	\subfigure[ {\scriptsize BDSD(17.23)}] {\includegraphics[width=.15\linewidth]{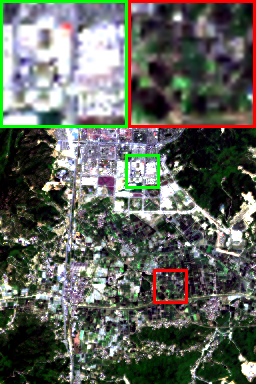}}    
	\subfigure[ {\scriptsize GS(14.21)}] {\includegraphics[width=.15\linewidth]{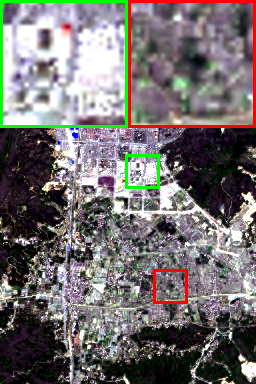}} 
	\subfigure[ {\scriptsize MIPSM(19.80)}] {\includegraphics[width=.15\linewidth]{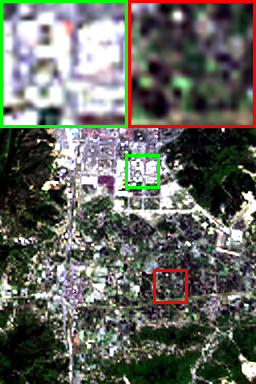}} 
	\subfigure[ {\scriptsize DRPNN(21.45)}] {\includegraphics[width=.15\linewidth]{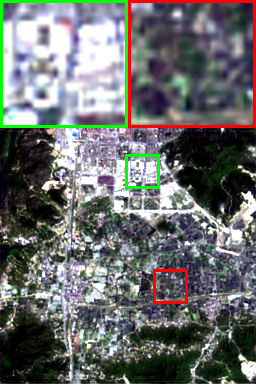}} 
	\subfigure[ {\scriptsize MSDCNN(21.94)}] {\includegraphics[width=.15\linewidth]{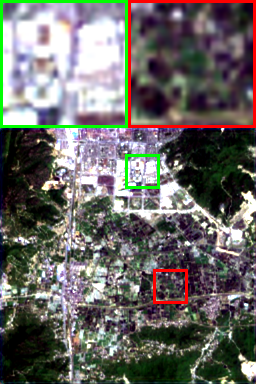}} 
	\subfigure[ {\scriptsize RSIFNN(17.14)}] {\includegraphics[width=.15\linewidth]{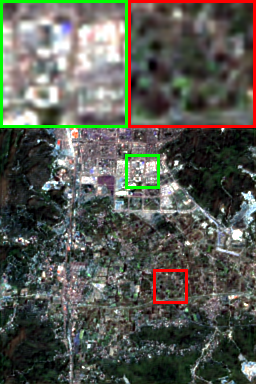}} 
	\subfigure[ {\scriptsize PANNET(20.09)}] {\includegraphics[width=.15\linewidth]{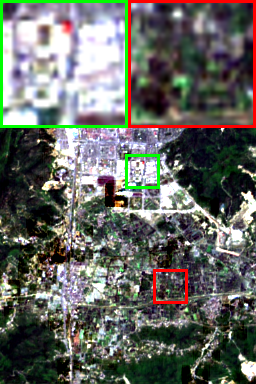}} 
	\subfigure[ {\scriptsize MHNet(19.58)}] {\includegraphics[width=.15\linewidth]{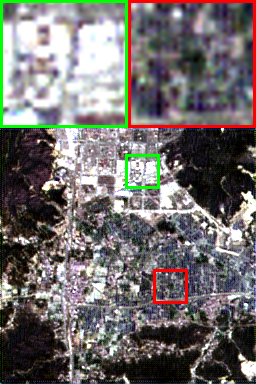}} 
	\subfigure[ {\scriptsize GPPNN(24.40)}] {\includegraphics[width=.15\linewidth]{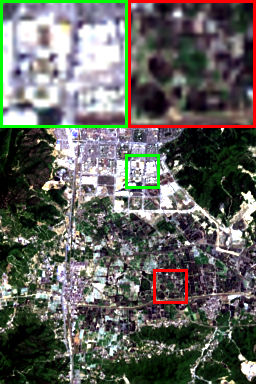}}    
	\caption{Visual inspection on Landsat8 dataset. The caption of each subimage displays the corresponding PSNR value.}
	\label{fig:L8}
\end{figure*}

\begin{table}[tbp]
	\centering
	\caption{The information of datasets. $B$ is the number of bands for multispectral images.
	}
	\resizebox{\linewidth}{!}{
		\begin{tabular}{lccc}
			\toprule
			& Landsat8 & GaoFen2 & QuickBird \\
			\midrule
			$B$ & 10    & 4     & 4 \\
			Resolution-MS & 256   & 256   & 256 \\
			Resolution-PAN & 512   & 1024  & 1024 \\
			\# Train/Val/Test & 350/50/100 & 350/50/100 & 474/103/100 \\
			\bottomrule
		\end{tabular}%
	}
	\label{tab:satellite}%
\end{table}%

\begin{table*}[htbp]
	\centering
	\caption{The PSNR values on validation datasets for GPPNN with different $K$ and $C$. The best value is highlighted by the \textbf{bold}. }
	\resizebox{\textwidth}{!}{
		\begin{tabular}{|c|ccccccc|ccccc|}
			\hline
			\multirow{2}[4]{*}{Satellites} & \multicolumn{7}{c|}{$K$ layers}                       & \multicolumn{5}{c|}{$C$ Filters} \bigstrut\\
			\cline{2-13}      & 2     & 4     & 6     & 8     & 10    & 12    & 14    & 8     & 16    & 32    & 64    & 128  \bigstrut\\
			\hline
			Landsat8 & 39.0648  & 39.5878  & 39.9876  & 40.0368  & \textbf{40.1336 } & 39.9531  & 40.0509  & 36.6455  & 39.6156  & 39.6702  & \textbf{40.0368 } & 39.0841  \bigstrut[t]\\
			QuickBird & 30.4994  & 30.4392  & 30.6370  & \textbf{30.5636 } & 30.4803  & 30.4773  & 30.5560  & 30.2962  & 30.4681  & 30.4592  & 30.5636  & \textbf{30.5979 } \\
			GaoFen2 & 36.7583  & 36.9740  & 36.2181  & \textbf{37.5606 } & 37.0589  & 36.7835  & 36.6840  & 35.8116  & 36.9061  & 36.2810  & \textbf{37.5606 } & 36.5873  \bigstrut[b]\\
			\hline
		\end{tabular}%
		
	}
	\label{tab:K_C}%
\end{table*}%

\begin{table*}[htbp]
	\centering
	\caption{The four metrics on test datasets. The best and the second best values are highlighted by the \textbf{bold} and \underline{underline}, respectively. The up or down arrow indicates higher or lower metric corresponds to better images.}
	\resizebox*{\textwidth}{!}{
		\begin{tabular}{|l|cccc|cccc|cccc|}
			\hline
			& \multicolumn{4}{c|}{Landsat8} & \multicolumn{4}{c|}{QuickBird} & \multicolumn{4}{c|}{GaoFen2} \bigstrut\\
			\cline{2-13}      & PSNR$\uparrow$ & SSIM$\uparrow$ & SAM$\downarrow$ & ERGAS$\downarrow$ & PSNR$\uparrow$ & SSIM$\uparrow$ & SAM$\downarrow$ & ERGAS$\downarrow$ & PSNR$\uparrow$ & SSIM$\uparrow$ & SAM$\downarrow$ & ERGAS$\downarrow$ \bigstrut\\
			\hline
			BDSD  & 33.8065  & 0.9128  & 0.0255  & 1.9128  & 23.5540  & 0.7156  & 0.0765  & 4.8874  & 30.2114  & 0.8732  & 0.0126  & 2.3963  \bigstrut[t]\\
			Brovey & 32.4030  & 0.8533  & 0.0206  & 1.9806  & 25.2744  & 0.7370  & 0.0640  & 4.2085  & 31.5901  & 0.9033  & 0.0110  & 2.2088  \\
			GS    & 32.0163  & 0.8687  & 0.0304  & 2.2119  & 26.0305  & 0.6829  & 0.0586  & 3.9498  & 30.4357  & 0.8836  & 0.0101  & 2.3075  \\
			HPF   & 32.6691  & 0.8712  & 0.0250  & 2.0669  & 25.9977  & 0.7378  & 0.0588  & 3.9452  & 30.4812  & 0.8848  & 0.0113  & 2.3311  \\
			IHS  & 32.8772  & 0.8615  & 0.0245  & 2.3128  & 24.3826  & 0.6742  & 0.0647  & 4.6208  & 30.4754  & 0.8639  & 0.0108  & 2.3546  \\
			Indusion & 30.8476  & 0.8168  & 0.0359  & 2.4216  & 25.7623  & 0.6377  & 0.0674  & 4.2514  & 30.5359  & 0.8849  & 0.0113  & 2.3457  \\
			SFIM  & 32.7207  & 0.8714  & 0.0248  & 2.0775  & 24.0351  & 0.6409  & 0.0739  & 4.8282  & 30.4021  & 0.8501  & 0.0129  & 2.3688  \\
			MIPSM & 35.4891  & 0.9389  & 0.0209  & 1.5769  & 27.7323  & 0.8411  & 0.0522  & 3.1550  & 32.1761  & 0.9392  & 0.0104  & 1.8830  \\
			DRPNN & 37.3639  & 0.9613  & 0.0173  & 1.3303  & 31.0415  & \underline{0.8993 } & 0.0378  & 2.2250  & \underline{35.1182 } & 0.9663  & 0.0098  & \underline{1.3078 } \\
			MSDCNN & 36.2536  & 0.9581  & 0.0176  & 1.4160  & 30.1245  & 0.8728  & 0.0434  & 2.5649  & 33.6715  & \underline{0.9685 } & 0.0090  & 1.4720  \\
			RSIFNN & 37.0782  & 0.9547  & 0.0172  & 1.3273  & 30.5769  & 0.8898  & 0.0405  & 2.3530  & 33.0588  & 0.9588  & 0.0112  & 1.5658  \\
			PANNET & \underline{38.0910 } & \underline{0.9647 } & \underline{0.0152 } & \underline{1.3021 } & 30.9631  & 0.8988  & \underline{0.0368 } & 2.2648  & 34.5774  & 0.9635  & \underline{0.0089 } & 1.4750  \\
			MHNet & 37.0049  & 0.9566  & 0.0189  & 1.3509  & \underline{31.1557 } & 0.8947  & \underline{0.0368 } & \underline{2.1931 } & 33.8930  & 0.9291  & 0.0176  & 1.3697  \\
			GPPNN & \textbf{38.9939 } & \textbf{0.9727 } & \textbf{0.0138 } & \textbf{1.2483 } & \textbf{31.4973 } & \textbf{0.9075 } & \textbf{0.0351 } & \textbf{2.1058 } & \textbf{35.9680 } & \textbf{0.9725 } & \textbf{0.0084 } & \textbf{1.2798 } \bigstrut[b]\\
			\hline
		\end{tabular}%
	}
	\label{tab:metric}%
\end{table*}%

\section{Experiments}
A series of experiments are carried out to evaluate the performance of GPPNN. SOTA deep learning based methods are selected for comparison, namely, MIPSM \cite{MIPSM}, DRPNN \cite{DRPNN}, MSDCNN \cite{MSDCNN}, RSIFNN \cite{RSIFNN}, PanNet \cite{PanNet}, and MHNet \cite{MHNet}. Our method is also compared with seven classic methods, including BDSD \cite{BDSD}, Brovey \cite{Brovey}, GS \cite{GS}, HPF \cite{HPF}, IHS fusion \cite{IHS}, Indusion \cite{Indusion}, SFIM \cite{SFIM}. The experiments are conducted on a computer with an Intel i7-9700K CPU at 3.60GHz and an NVIDIA GeForce RTX 2080ti GPU.

\begin{figure*}[ht]
	\centering
	\subfigure[ {\scriptsize LRMS}] {\includegraphics[width=.15\linewidth]{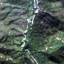}}  
	\subfigure[ {\scriptsize PAN}] {\includegraphics[width=.15\linewidth]{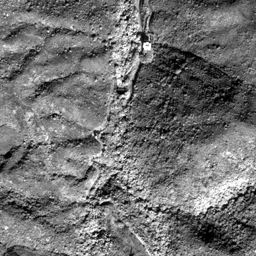}}  
	\subfigure[ {\scriptsize GT(PSNR)}] {\includegraphics[width=.15\linewidth]{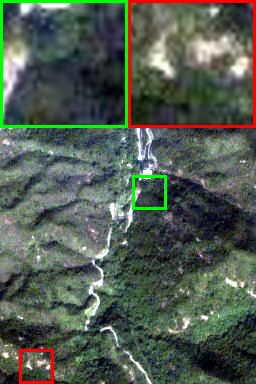}}  
	\subfigure[ {\scriptsize BDSD(16.23)}] {\includegraphics[width=.15\linewidth]{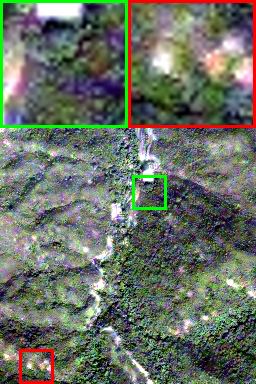}}    
	\subfigure[ {\scriptsize GS(15.80)}] {\includegraphics[width=.15\linewidth]{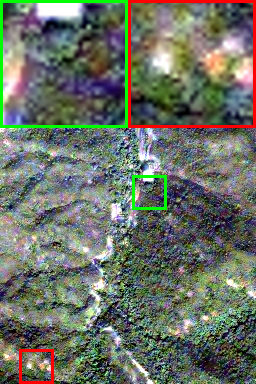}} 
	\subfigure[ {\scriptsize MIPSM(16.84)}] {\includegraphics[width=.15\linewidth]{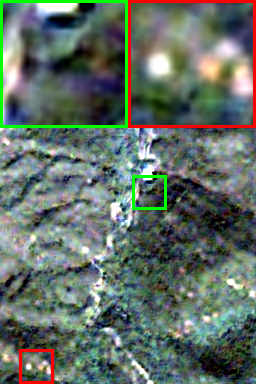}} 
	\subfigure[ {\scriptsize DRPNN(18.85)}] {\includegraphics[width=.15\linewidth]{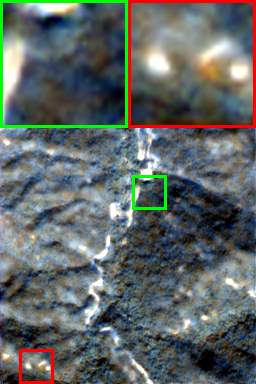}} 
	\subfigure[ {\scriptsize MSDCNN(18.15)}] {\includegraphics[width=.15\linewidth]{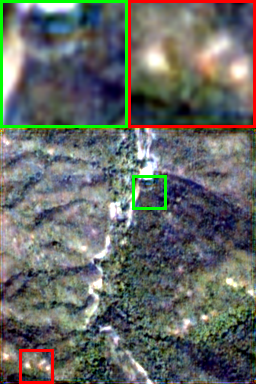}} 
	\subfigure[ {\scriptsize RSIFNN(17.33)}] {\includegraphics[width=.15\linewidth]{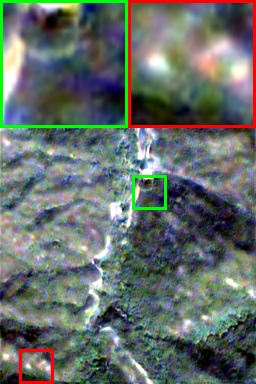}} 
	\subfigure[ {\scriptsize PANNET(18.22)}] {\includegraphics[width=.15\linewidth]{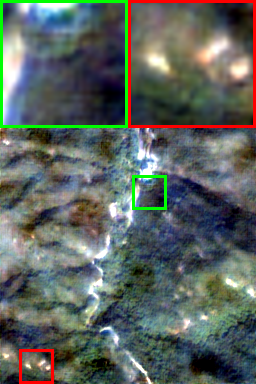}} 
	\subfigure[ {\scriptsize MHNet(19.50)}] {\includegraphics[width=.15\linewidth]{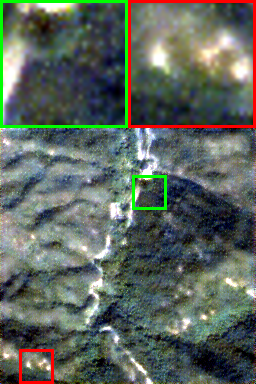}} 
	\subfigure[ {\scriptsize GPPNN(21.13)}] {\includegraphics[width=.15\linewidth]{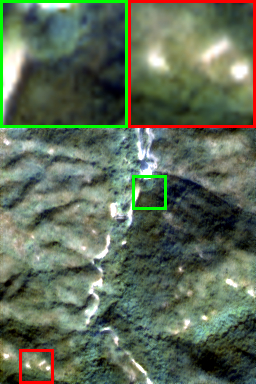}}  
	\caption{Visual inspection on QuickBird dataset. The caption of each subimage displays the corresponding PSNR value.}
	\label{fig:QB}
\end{figure*}

\subsection{Datasets and metrics}
Remote sensing images acquired by three satellites are used in our experiments, including Landsat8, QuickBird and GaoFen2, the basic information of which is listed in Table \ref{tab:satellite}. For each satellite, we have hundreds of image pairs, and they are divided into three parts for training, validation and test. Note that we determine $K$ and $C$ on the validation dataset. In the training set, the multispectral images are cropped into patches with the size of $32\times 32$, and the corresponding PAN patches are with the size of $64\times 64$ (for Landsat8) or $128\times 128$ (for GaoFen2 and QuickBird). For numerical stability, each patch is normalized by dividing the maximum value to make the pixels range from 0 to 1. 

Four popular metrics are used to evaluate the algorithms' performances, including peak signal-to-noise ratio (PSNR), structural similarity (SSIM) and erreur relative globale adimensionnelle de synthese (ERGAS) and spectral angle mapper (SAM). The first three metrics measure the spatial distortion, and the last one measures the spectral distortion. An image is better if its PSNR and SSIM are higher, and ERGAS and SAM are lower. 

\subsection{The effect of depth and width}\label{sec:K_C}
The network's depth $K$ and width $C$ play significant roles. Table \ref{tab:K_C} lists the PSNR values on validation datasets for GPPNN with different $K$ and $C$. At first, $C$ is fixed to 64, and $K$ is set to 2, 4, $\cdots$, 14. It is shown that more layers do not necessarily increase the PSNR value, and $K=8$ strikes the balance between performance and the number of weights. The reason may be that it is not easy to train a GPPNN with more layers. Then, we fix $K$ to 8 and set $C$ to 8, 16, 32, 64 and 128. The similar conclusion can be drawn, and the best choice for $C$ is 64. In summary, our GPPNN is configured with $K=8$ layers and $C=64$ filters in the next experiments.

\subsection{Comparison with SOTA methods}
\begin{figure*}[t]
	\centering
	\subfigure[ {\scriptsize LRMS}] {\includegraphics[width=.15\linewidth]{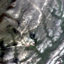}}  
	\subfigure[ {\scriptsize PAN}] {\includegraphics[width=.15\linewidth]{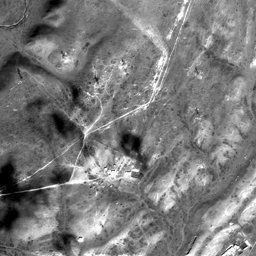}}  
	\subfigure[ {\scriptsize GT(PSNR)}] {\includegraphics[width=.15\linewidth]{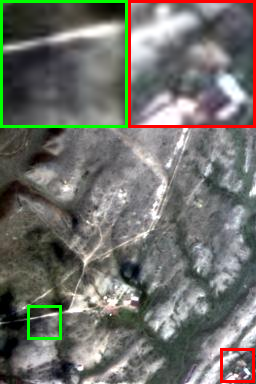}}  
	\subfigure[ {\scriptsize BDSD(21.86)}] {\includegraphics[width=.15\linewidth]{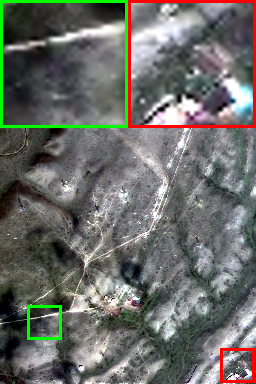}}    
	\subfigure[ {\scriptsize GS(21.43)}] {\includegraphics[width=.15\linewidth]{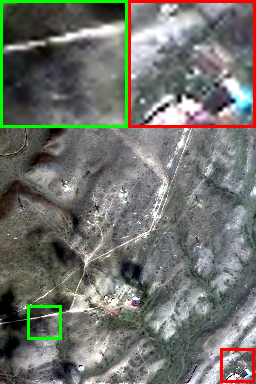}} 
	\subfigure[ {\scriptsize MIPSM(22.11)}] {\includegraphics[width=.15\linewidth]{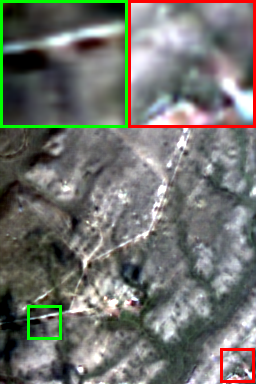}} 
	\subfigure[ {\scriptsize DRPNN(26.41)}] {\includegraphics[width=.15\linewidth]{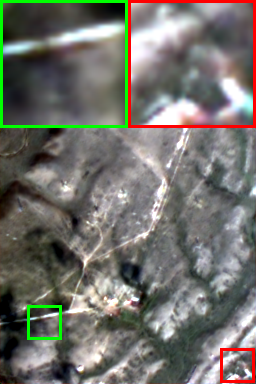}} 
	\subfigure[ {\scriptsize MSDCNN(26.90)}] {\includegraphics[width=.15\linewidth]{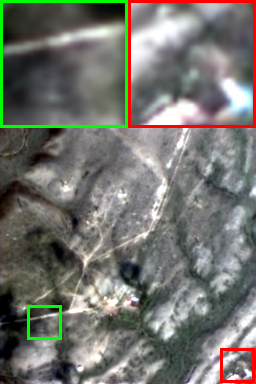}} 
	\subfigure[ {\scriptsize RSIFNN(24.74)}] {\includegraphics[width=.15\linewidth]{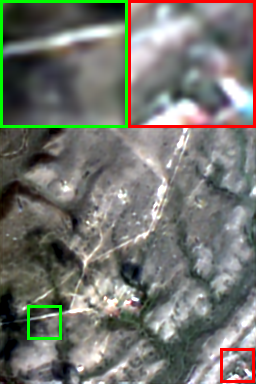}} 
	\subfigure[ {\scriptsize PANNET(26.72)}] {\includegraphics[width=.15\linewidth]{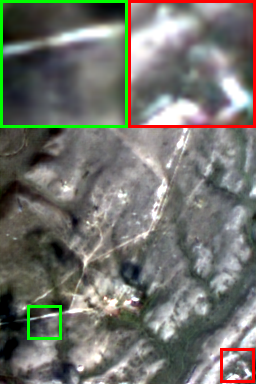}} 
	\subfigure[ {\scriptsize MHNet(25.17)}] {\includegraphics[width=.15\linewidth]{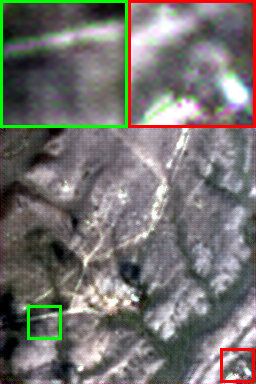}} 
	\subfigure[ {\scriptsize GPPNN(28.70)}] {\includegraphics[width=.15\linewidth]{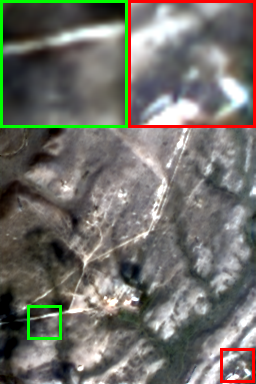}}   
	\caption{Visual inspection on GaoFen2 dataset. The caption of each subimage displays the corresponding PSNR value.}
	\label{fig:GF}
\end{figure*}

The evaluation metrics on three datasets are reported in Table \ref{tab:metric}. It is found that GPPNN outperforms other methods regarding all metrics on three satellites. Figs. \ref{fig:L8}, \ref{fig:QB} and \ref{fig:GF} show the RGB bands of the three satellites for some representative methods. Our GPPNN is the closest to the ground truth. From the amplified local regions in Fig. \ref{fig:L8}, it found that BDSD, GS, MIPSM, DRPNN, PANNET suffer from spatial distortion, and GS, MSDCNN, RSIFNN and MHNet suffer from spectral distortion. However, our GPPNN has the smallest spatial and spectral distortions. As for Fig. \ref{fig:QB}, it is a difficult case. It is shown that the most methods have obvious artifacts or noise, and their images are blurring or spectrally distorted. Our GPPNN is without artifacts, noise or spectral distortion. As shown in Fig. \ref{fig:GF}, it is observed that compared with other methods, our GPPNN has finer-grained textures and coarser-grained structures. 

\begin{table*}[htbp]
	\centering
	\caption{The results of ablation experiments on the Landsat8 dataset.}
	{\footnotesize 
		\begin{tabular}{|c|cccc|cccc|}
			\hline
			\multirow{2}[2]{*}{Configurations} & Proximal & Sharing & Block for & Transpose- & \multirow{2}[2]{*}{PSNR$\uparrow$} & \multirow{2}[2]{*}{SSIM$\uparrow$} & \multirow{2}[2]{*}{SAM$\downarrow$} & \multirow{2}[2]{*}{ERGAS$\downarrow$} \bigstrut[t]\\
			& Module & Weights & Eq. (\ref{eq:naive_model}) & ment  &       &       &       &  \bigstrut[b]\\
			\hline
			I     & \texttimes & \texttimes & \texttimes & \texttimes & 37.0404  & 0.9498  & 0.0180  & 1.4246  \bigstrut[t]\\
			II    & $\checkmark$ & $\checkmark$ & \texttimes & \texttimes & 38.1650  & 0.9669  & 0.0164  & 1.2943  \\
			III   & $\checkmark$ & \texttimes & $\checkmark$ & \texttimes & 38.3213  & 0.9682  & 0.0155  & 1.3215  \\
			IV    & $\checkmark$ & \texttimes & \texttimes & $\checkmark$ & 38.5487  & 0.9700  & 0.0150  & 1.2746  \\  \hline
			GPPNN & $\checkmark$ & \texttimes & \texttimes & \texttimes & \textbf{38.9939 } & \textbf{0.9727 } & \textbf{0.0138 } & \textbf{1.2483 } \bigstrut[t]\\
			\hline
		\end{tabular}%
	}
	\label{tab:ablation}%
\end{table*}%

\subsection{Ablation experiments}\label{sec:ablation_exp}
To further investigate the role of some modules in the proposed GPPNN, a series of ablation experiments are carried out. There are 5 different configurations and the results of ablation experiments are shown in Table \ref{tab:ablation}. 

(I) The proximal operators make the current HRMS image restricted to deep priors. In the first experiment, we delete proximal modules (namely, the convolutional units in Eqs. (\ref{eq:nn_LR4})\&(\ref{eq:nn_PAN4})) to verify the necessity of deep priors. Table \ref{tab:ablation} shows that deleting proximal modules make all metrics dramatically get worse. Therefore, the deep prior plays a significant role in our network.

(II) In the second experiment, we share the weights of all layers. In other words, the network contains only an MS Block and a PAN block, and the network is repeatedly fed with the current HRMS image $K$ times. The results in Table \ref{tab:ablation} demonstrate that sharing the weights will weaken our network's performance. 

(III) As illustrated in Section \ref{sec:model_formulation}, the original problem Eq. (\ref{eq:naive_model}) is split into an LRMS-aware subproblem and a PAN-aware subproblem. Now, to verify the rationality, we generalize Eq. (\ref{eq:naive_model}) as a neural network with the same techniques for GPPNN. 
We exploit this block to build a neural network corresponding to Eq. (\ref{eq:naive_model}). From Table \ref{tab:ablation}, we learn that the network for Eq. (\ref{eq:naive_model}) is worse than GPPNN. It is necessary to consider two deep priors to separately account for the generative models of LRMS and PAN images.

(IV) In the last experiment, the convolutional kernel in Eq. (\ref{eq:nn_LR3})/(\ref{eq:nn_PAN3}) is replaced by the kernel in Eq. (\ref{eq:nn_LR1})/(\ref{eq:nn_PAN1}) with the rotation of 180\textdegree\ to force them to satisfy the transposing requirement. It is found that, if the two kernels transpose to each other, the metrics will slightly become worse. The reason may be that the model with transposed kernels has fewer degrees of freedom weakening network's performance.

\section{Conclusion and Future Work}
This paper provides a new paradigm combining deep unrolling and observation models of pan-sharpening. We develop a model-driven pan-sharpening network, GPPNN, by alternatively stacking MS and PAN blocks whose designs are inspired by two optimization problems.  Experiments on three satellites show that our network outperforms SOTA methods. And a series of ablation experiments verify the rationality of our network structure. 

Remark that each satellite has its unique imaging parameters, including $\bm{D}, \bm{K}$ and $\bm{S}$. GPPNN trained on a satellite cannot be generalized to another satellite. Hence, the future work is how to improve the generalization of GPPNN.

\section*{Acknowledgement}
This work was supported
in part by the National Key Research and Development Program of China
under Grant 2018AAA0102201, and in part by the National Natural Science
Foundation of China under Grants 61976174, 61877049.

{\small
\bibliographystyle{ieee_fullname}
\bibliography{egbib}
}

\end{document}